\documentclass{article}

\usepackage[preprint]{nips_2018}

\usepackage[utf8]{inputenc} % allow utf-8 input
\usepackage[T1]{fontenc}    % use 8-bit T1 fonts
\usepackage{hyperref}       % hyperlinks
\usepackage{url}            % simple URL typesetting
\usepackage{booktabs}       % professional-quality tables
\usepackage{amsfonts}       % blackboard math symbols
\usepackage{microtype}      % microtypography
\usepackage{times}
\usepackage{xcolor}
\usepackage{soul}
\usepackage[small]{caption}
\usepackage{graphicx}
\usepackage{amsmath}
\usepackage{makecell}
\usepackage{threeparttable}
\usepackage{booktabs}
\usepackage[normalem]{ulem}
\usepackage{multirow}
\title{SGAD: Soft-Guided Adaptively-Dropped Neural Network}

\author{
  Zhisheng Wang$^1$$^*$, Fangxuan Sun$^1$$^*$, Jun Lin$^1$, Zhongfeng Wang$^1$ and Bo Yuan$^2$ \\
  $^1$ School of Electronic Science and Engineering, Nanjing University, P.R. China\\
  $^2$ Department of Electrical Engineering, City University of New York, City College\\
  \texttt{\{zswang, fxsun\}@smail.nju.edu.cn}  \\
  \texttt{\{jlin, zfwang\}@nju.edu.cn, byuan@ccny.cuny.edu} \\
}

\begin{document}

\maketitle
\footnotetext[1]{Authors contributed equally.}
\begin{abstract}
	Deep neural networks (DNNs) have been proven to have many redundancies. Hence, many efforts have been made to compress DNNs. However, the existing model compression methods treat all the input samples equally while ignoring the fact that the difficulties of various input samples being correctly classified are different. To address this problem, DNNs with adaptive dropping mechanism are well explored in this work. To inform the DNNs how difficult the input samples can be classified, a guideline that contains the information of input samples is introduced to improve the performance. Based on the developed guideline and adaptive dropping mechanism, an innovative soft-guided adaptively-dropped (SGAD) neural network is proposed in this paper. Compared with the 32 layers residual neural networks, the presented SGAD can reduce the FLOPs by \(77\%\) with less than \(1\%\) drop in accuracy on CIFAR-10.%, which is superior to all existing works.
\end{abstract}

\section{Introduction}\label{sec: intro}
Deep neural networks (DNNs) have achieved the state-of-the-art accuracy  and gained wide adoption in various artificial intelligence (AI) fields, such as computer vision, speech recognition and nature langue processing~\cite{He2016Deep,krizhevsky2012imagenet,simonyan2014very,amodei2015deep}. However, the remarkable accuracy of DNNs comes at the expense of huge computational cost, which has already posed severe challenges on the existing DNN computing hardware performance in terms of processing time and power consumption. Even worse, it is widely acknowledged that the computational cost of modern DNNs will continue to increase rapidly due to the ever-growing demands for improved accuracy in AI applications. Consider the limited progress of hardware technology, the huge computational cost of DNNs, if not being properly addressed, would largely prevent the large-scale deployment of DNNs on various resource-constrained platforms, such as mobile devices and Internet-of-Thing (IoT) equipment. 

To address this challenge, several computation-reducing approaches have been proposed in~\cite{wen2016learning,Wen2017Coordinating,han2016deep,sun2016intra,Garipov2016Ultimate}. To date, most of the existing works focus on modifying the original popular DNN architectures via different techniques (such as pruning and decomposition etc.). In those model-pruning/decomposition works, all the input samples are treated equally and they are processed by all the layers of DNNs. Consider that shallow models with relatively poor model capacities can also correctly classify some input samples, thus, different samples in the same dataset exhibit different levels of ease on accurate classification. By leveraging such characteristics, an input-specific adaptive computational approach can be exploited to avoid unnecessary computation. 

A natural way to skip a layer is to add a bypass which directly outputs the inputs. Among various DNNs, residual networks~\cite{He2016Deep} (ResNets) exhibit a unique architecture which is friendly to the adaptive computational approach. Hence, this paper focuses on the adaptive computation of ResNets. There are two more reasons for choosing ResNets,  1) ResNet is the currently most popular and widely deployed DNN architecture, especially in computer vision field; 2) previous work~\cite{veit2016residual} showed that ResNets can be seen as \emph{ensembles} of many shallow blocks with weak dependencies which can be utilized for adaptive computation.

\begin{figure} [hbt]
	\centering
	\includegraphics[width=5.6in]{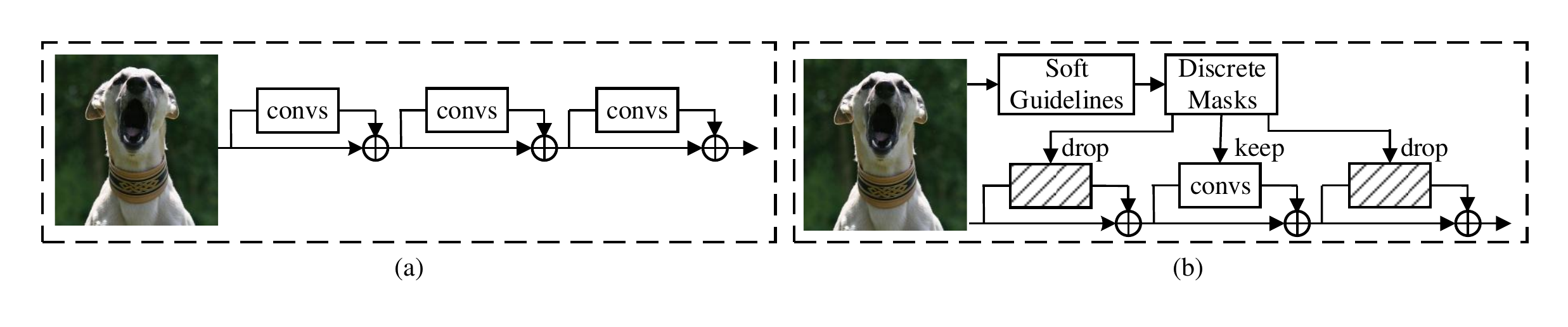}
	\caption{Overview of adaptive dropped mechanism. Blocks denote the residual blocks that consist of several convolutional layers. (a) ResNets without any modifications. (b) ResNets with adaptive dropped mechanism. The dropped blocks are denoted by those striated blocks.}\label{fig: intro}
\end{figure}

In this paper, we propose a novel end-to-end trainable soft-guided adaptively-dropped neural network (SGAD) to reduce the input-specific redundant computations while retaining high accuracy. As shown in Fig.~\ref{fig: intro}, all blocks in the original ResNet are always busy. However, in SGAD, each block will be adaptively dropped according to the input samples. To smartly and efficiently decide which blocks will be dropped, a soft guideline is developed to generate a group of discrete masks. Experimental results show that the proposed SGAD can reduce \(77\%\) floating-point operations (FLOPs) with less than \(1\%\) accuracy loss compared with ResNet-32 on CIFAR-10. On CIFAR-100, SGAD can improve the accuracy by \(0.47\%\) with \(23\%\) less FLOPs as compared with ResNet-32. The contributions of this paper are summarized as follows:
\begin{itemize}
	\item A novel soft information-based guideline is proposed to quantize the level of difficulties of input samples being classified correctly. Such guideline is then used to direct the expected drop ratio of residual blocks during training via an efficient mapping strategy. At the inference stage, the guideline can be removed without incurring additional overhead.
	\item We introduce a small but efficient model with binary output, which determines the positions of layers that will be skipped according to the current input sample under the direction of the proposed guideline. Straight through estimator (STE)~\cite{Bengio2013Estimating} is introduced to approximate the non-differential of the rounding function during the training phase.
	\item The learned dropping behavior of SGAD is explored. Our experiments show that layers of original network (e.g. ResNet-32) with less contribution to the model capacity are likely to be dropped in SGAD-based model (e.g. SGAD-32).
\end{itemize}

\section{Related Works}
The proposed SGAD is motivated by recent studies on exploring the behavior of residual networks. Andreas~\textit{et al.} found that residual networks can be seen as ensembles of many weakly-dependent paths with varying lengths, where only the short paths are needed during training~\cite{veit2016residual}. Besides, removing individual layers from a trained residual network at test time only leads to misclassification on few \textit{borderline} samples with minor accuracy drop~\cite{greff2016highway}. These observations indicate that most input samples may be easily classified with limited number of layers, thus we can adaptively allocate different computation budgets between ``easy'' and ``hard'' samples. Several approaches have been proposed based on this concept. 

\textbf{The early-termination approaches}~\cite{bolukbasi2017adaptive,teerapittayanon2016branchynet,panda2016conditional} add additional side-branch classifiers inside a deep neural network. Hence, input samples that are judged as being able to be classified by a certain side-branch classifier can exit from the network immediately without executing the whole model. In contrast, the proposed SGAD enables adaptive-computation behavior by utilizing the ensemble nature of residual networks. Neither hand-crafted network architectures nor extra side-branch classifier is needed, thereby making our approach more simple and effective.

\textbf{The adaptive computation approaches} are close to our work. Spatially Adaptive Computation Time (SACT)~\cite{figurnov2016spatially} dynamically decides the number of executed layers inside a set of residual units. SkipNet~\cite{wang2017skipnet} and BlockDrop~\cite{wu2017blockdrop} utilize reinforcement learning to dynamically choose executed residual units in a pretrained ResNet for different input samples. Adanets~\cite{veit2017convolutional} enable adaptive computation graphs by adding layer-wise gating functions to decide whether to skip the computation of a certain layer or not. Different from these approaches, the proposed SGAD uses a shallow network, whose behavior is guided by an extra guideline during training, to generate binary vectors for adaptively masking those unused residual units. Compared to above mentioned approaches, SGAD is able to achieve higher savings in computational cost with no accuracy loss in most cases.

\section{Soft-Guided Adaptively-Dropped Approach}
In this section, we present the soft-guided adaptively-dropped neural network. First, we introduce a binary mask network (BMNet) to decide which blocks should be used for a specific input. The size of BMNet is quite small and hence it introduces very little computation overhead. Then, in order to solve the non-differentiable problem incurred by using these discrete binary masks in the training phase, straight through estimator (STE)~\cite{Bengio2013Estimating} is introduced to approximate the gradient of the original non-differential rounding function during back propagation. Finally, we propose a soft guideline network (SGNet) to improve the overall classification accuracy. The SGNet can extract the soft information of different inputs during the training phase, and thereby aiding the training of BMNet through a regularization term to force BMNet drop dynamically. At the inference phase, the regularization term is no longer used, thus the SGNet can be removed.
\begin{figure*} [hbt]
	\centering
	\includegraphics[width=5.5in]{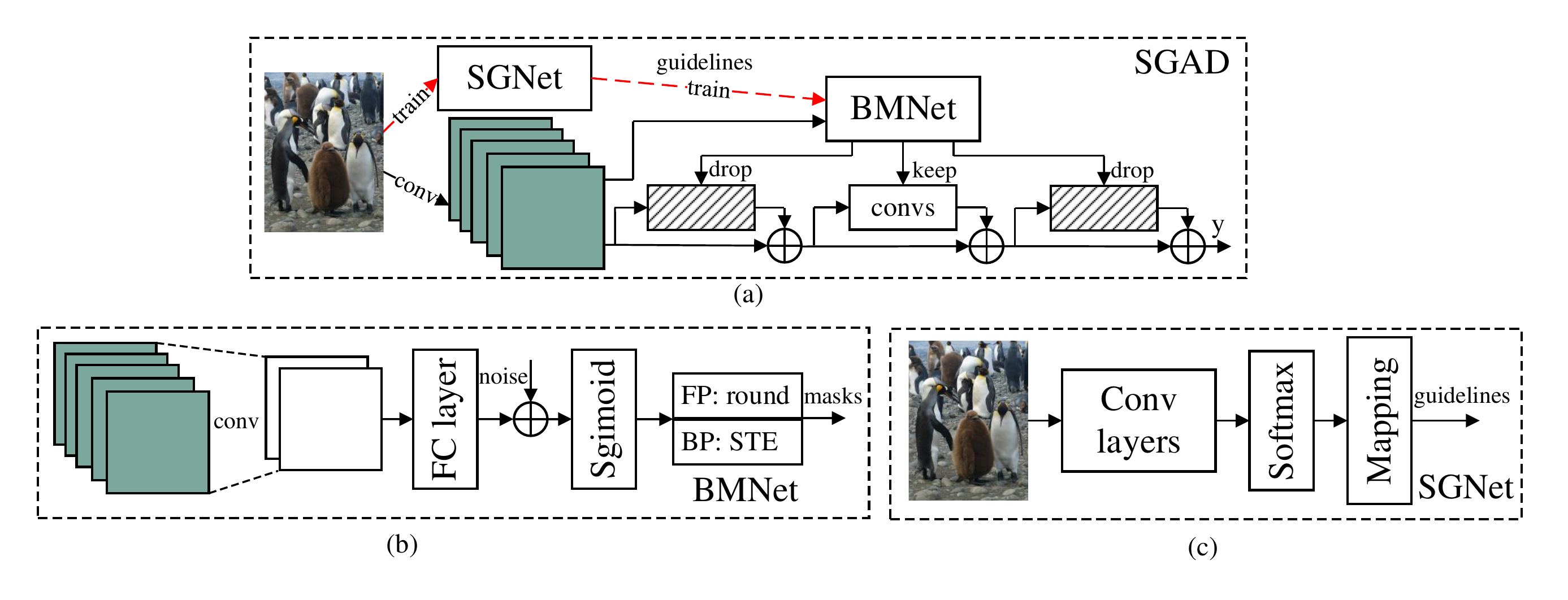}
	\caption{Architecture of the proposed SGAD. (a). The overview of the presented algorithm. The activations after the first single convolutional layer of ResNet is denoted by green blocks. The red dot lines mean that the SGNet will  be active only in the training phase. (b) Architecture of BMNet. The binary rounding and sinc function-based estimator are used in the phase of forward and backward propagation, respectively. (c). Architecture of SGNet. The block named \emph{Conv layers} denotes a group of convolutional layers which can be flexibly adjusted.}\label{fig: SGAD}
\end{figure*}
\subsection{Binary Mask}\label{sec: mask}
Generally, the main part of ResNets consists of several blocks. Let \(\boldsymbol{z}^{i}\) and \(\boldsymbol{z}^{i+1}\) be the input and output of the (\(i\)+1)-th block, respectively. The computation of \(\boldsymbol{z}^{i+1}\) is shown below:
\begin{eqnarray}\label{eqn: block}
\begin{split}
\boldsymbol{z}^{i+1}=\boldsymbol{z}^{i}+f(\boldsymbol{z}^{i})
\end{split},
\end{eqnarray}
where the details of \(f(\boldsymbol{z})\) can be referred in~\cite{He2016Deep}. Beside the blocks, the ResNets usually start with a single convolutional layer and end with a fully-connected layer.

\textbf{Binary Mask: } As indicated in the first paragraph of this section, we introduce a binary mask to determine whether each block should be skipped or not in the inference of a specific input sample. Specifically, for the \(i\)-th block, its output determined by the binary mask is as follows:
\begin{eqnarray}\label{eqn: mask}
\boldsymbol{z}^{i+1}=
\begin{cases}
\boldsymbol{z}^{i}+f(\boldsymbol{z}^{i})& \text{$m(z^{1}; \theta_{m})=1$}\\
\boldsymbol{z}^{i}& \text{$m(z^{1}; \theta_{m})=0$}
\end{cases},
\end{eqnarray}	
where \(z^{1}\) denotes the input of the first block, namely the output of the single convolutional layer in ResNet. \(m(\boldsymbol{z}^{1}; \theta_{m})\) is a hypothesis with weight \(\theta_{m}\) which decides whether or not this block should be dropped. To simplify the deduction of gradient, Eq.~(\ref{eqn: mask}) can be rewritten in the following form:
\begin{eqnarray}\label{eqn: mask2}
\begin{split}
\boldsymbol{z}^{i+1}=\boldsymbol{z}^{i}+m(\boldsymbol{z}^{1}; \theta_{m})  f(\boldsymbol{z}^{i})
\end{split}.
\end{eqnarray}

Assume that a batch of data contains \(N\) pairs {\((\boldsymbol{x}_{1}, \boldsymbol{y}_{1}),\dots,(\boldsymbol{x}_{N}, \boldsymbol{y}_{N})\)} during the training phase. The weights of the \(i\)-th block of ResNets are denoted by \(\theta^{i}\). The update of \(\theta^{i}\) at the (\(t\)+\(1\))-th iteration \(\theta_{t+1}^{i}\) can be written as belows:
\begin{eqnarray}\label{eqn: loss2}
\begin{split}
\theta_{t+1}^{i} = \theta_{t}^{i} - lr \times \frac{\partial R}{\partial \theta_{t}^{i}}
\end{split},
\end{eqnarray}
where \(lr\) and \(R\) denote the learning rate used in the training phase and the training loss at the \(t\)-th iteration, respectively. 

For original ResNets, the gradient \(\frac{\partial R}{\partial \theta_{t}^{i}}\) can be represented as follows:
\begin{eqnarray}\label{eqn: loss3}
\begin{split}
&\frac{\partial R}{\partial \theta_{t}^{i}} = \frac{1}{N}\sum_{n=1}^{N}\frac{\partial R_{n}}{\partial \boldsymbol{z}_{n}^{L}}\frac{\partial \boldsymbol{z}_{n}^{L}}{\partial \boldsymbol{z}_{n}^{L-1}} \dots \frac{\partial \boldsymbol{z}_{n}^{i+1}}{\partial \theta_{t}^{i}} = \frac{1}{N}\sum_{n=1}^{N} \frac{\partial R_{n}}{\partial \boldsymbol{z}_{n}^{L}} (\prod_{j=i+1}^{L-1} \frac{\partial \boldsymbol{z}_{n}^{j+1}}{\partial \boldsymbol{z}_{n}^{j}}) \frac{\partial \boldsymbol{z}_{n}^{i+1}}{\partial \theta_{t}^{i}} \\
&=\frac{1}{N}\sum_{n=1}^{N} \frac{\partial R_{n}}{\partial \boldsymbol{z}_{n}^{L}} (\prod_{j=i+1}^{L-1}(1+\bigtriangledown f^{j}_{n}(\boldsymbol{z}^{j}_{n}))) \bigtriangledown f^{i}_{n}(\theta^{i}_{t})
\end{split},
\end{eqnarray}
where \(L\) denotes the number of blocks in a ResNet. \(\bigtriangledown f^{j}_{n}(\boldsymbol{z}^{j}_{n})\) denotes the differential of \(\boldsymbol{z}_{n}^{j+1}\) to \(\boldsymbol{z}_{n}^{j}\) and \( \bigtriangledown f^{i}_{n}(\theta^{i}_{t})\) denotes the differential of \(\boldsymbol{z}_{n}^{i+1}\) to \(\theta_{t}^{i}\). Taking the binary mask into consideration, the update of gradients can be represented as:
\begin{eqnarray}\label{eqn: loss4}
\begin{split}
&\frac{\partial R^{'}}{\partial \theta_{t}^{i}} = \frac{1}{N}\sum_{n=1}^{N} \frac{\partial R_{n}}{\partial \boldsymbol{z}_{n}^{L}}(\prod_{j=i+1}^{L-1}(1+m_{n}^{j}(\boldsymbol{z}^{1}_{n}; \theta_{m})\bigtriangledown f^{j}_{n}(\boldsymbol{z}^{j}_{n}))) (m_{n}^{i}(\boldsymbol{z}^{1}_{n};  \theta_{m})\bigtriangledown f^{i}_{n}(\theta^{i}_{t}))
\end{split}.
\end{eqnarray}

Generally, the gradients calculated in the training phase is much less than 1 (the magnitude of gradients are about \(10^{-6}\) according to Section.\ref{sec: dis}). Hence,  \((\prod_{j=i+1}^{L-1}(1+m_{n}^{j}(\boldsymbol{z}^{1}_{n}; \theta_{m})\bigtriangledown f^{j}_{n}(\boldsymbol{z}^{j}_{n})))\) in Eq.~(\ref{eqn: loss3}) can be seen as 1. Eq.~(\ref{eqn: loss3}) can thusly be simplified to:
\begin{eqnarray}\label{eqn: loss5}
\begin{split}
\frac{\partial R^{'}}{\partial \theta_{t}^{i}} \approx \frac{1}{N}\sum_{n=1}^{N} (m_{n}^{i}(\boldsymbol{z}^{1}_{n};  \theta_{m})  \frac{\partial R_{n}}{\partial \boldsymbol{z}_{n}^{L}} \bigtriangledown f^{i}_{n}(\theta^{i}_{t}))
\end{split}.
\end{eqnarray} 

Let \(rats_{b}^{i}=\frac{\sum_{n=1}^{N} m_{n}^{i}(\boldsymbol{z}^{i}_{n}; \theta_{m})}{N}\) be the ratio that does not dropped in a batch for the \(i\)-th block. Combining Eqs.~(\ref{eqn: loss3} - \ref{eqn: loss5}) and the definition of \(rats_{b}^{i}\), the updating of the weights of ResNets with binary mask can be approximated as follows:
\begin{eqnarray}\label{eqn: loss7}
\begin{split}
\theta_{t+1}^{i} \sim \theta_{t}^{i} - (lr \times rats_{b}^{i})  \frac{\partial R}{\partial \theta_{t}^{i}}
\end{split},
\end{eqnarray}
where \(lr \times rats_{b}^{i}\) is the actual learning rate. Since \(rats_{b}^{i}\) in different blocks are not the same, each block will have an unique learning rate. Hence, the proposed binary mask can adaptively adjust the learning rate of different blocks according to the \emph{level of contributions to the model capacity (LCMC)} of blocks~\cite{veit2016residual}. To explore which blocks contribute more to the model capacity, we will study the dropping behavior of SGAD with details discussed in Section~\ref{sec: dis}.

The design of binary mask called BMNet is introduced and shown in Fig.~\ref{fig: SGAD} (b). Note that small perturbations can result in quite different binary masks if the output of the sigmoid unit is near the rounding threshold (0.5), thereby making the BMNet instable. Inspired by \cite{salakhutdinov2009semantic}, additive noises are injected before the sigmoid unit. The magnitude of noise is increased over time so that the magnitude of inputs will also be increased to alleviate impact of the noise. With the use of this method, the sigmoid unit can be trained to be saturated in nearly 0 or 1 for all input samples. Hence, more stable and confident decisions are generated during both the training and the inference phases.

\subsection{Soft Guideline}\label{sec: sg}
With the proposed BMNet, an adaptively dropped ResNet can be realized. However, how BMNet decides the dropping ratio is unknown. Consider that our goal is to make the networks adaptively adjust the computational complexity according to the difficulty of classification of input samples, the information of whether or not the input samples are easy to be classified should be generated and sent to BMNet to improve the correctness of decisions. Based on this concept, an additional network, called the soft guideline network (SGNet), is proposed to produce the required information and guide the dropping behavior of the BMNet.

\textbf{Soft Guideline: } Generally, each input sample couples with a \emph{hard target} which only contains the information of the truth label class. The information of whether or not the input samples are easy to be classified can not be gained from the \emph{hard targets}. Inspired by ~\cite{hinton2015distilling}, the \emph{soft target}, namely the class probabilities produced by the softmax layer, can provide much more information than the \emph{hard target}. In this paper, the \emph{soft target} of the SGNet is used to obtain the information which indicates the difficulty of classification. More specifically, the variance of the \emph{soft target} is used as the guideline. For input sample \(\boldsymbol{x}_{n}\) (\(n \in \{1,2,\cdots,N\}\)), corresponding variance \(var^n\) can be written as follows:
\begin{eqnarray}\label{eqn: soft1}
\begin{split}
var^n=&\frac{1}{M}\sum_{i=1}^{M}[(\frac{e^{s^n_{i}}}{\sum_{j} e^{s^n_{j}}} - \frac{1}{M})^{2}] = \frac{\sum_{i}e^{2s^n_{i}}}{M(\sum_{j}e^{s^n_{j}})^{2}} - \frac{1}{M^{2}} \in [0, \frac{1}{M})
\end{split},
\end{eqnarray}
where \(M\) is the number of classes. \(s^n_{1}, s^n_{2}, \dots, s^n_{M}\) are the elements of the softmax output for \(\boldsymbol{x}_{n}\). Intuitively, smaller value of \(var^n\) indicates that the SGNet is less confident for its classification result. Thus, it tends harder to correctly classify \(\boldsymbol{x}_{n}\).

In order to make the BMNet learn to adaptively drop more (less) residual blocks for easily (hardly) classified input samples, the guideline \(var^n\) is first transformed to produce an expected drop ratio \(rat^n_s \in [0,1]\). Easily classified \(\boldsymbol{x}_{n}\) will have higher \(rat^n_s\). Then, the L1-norm between the \(rat^n_s\) and the calculated drop ratio for all input samples in a batch, denoted as \(R^{m}\), is added to the loss function as an regularizer to push the BMNet allocate desired drop ratio for different input samples, where
\begin{eqnarray}\label{eqn: soft4}	
R^{m}=\frac{1}{N} \sum_{n=1}^{N} \parallel rat^n_{s}- (1-\frac{1}{L}\sum_{j=1}^{L}m^n_{j}(\boldsymbol{z}^{1}_n; \theta_{m})) \parallel_{1},
\end{eqnarray}
where \(1-\frac{1}{L}\sum_{j=1}^{L}m^n_{j}(\boldsymbol{z}^{1}_n; \theta_{m})\) is the measured average drop ratio (computed by BMNet). The application of this regularizer can push the actual drop ratio and the desired drop ratio closer.

Based on the above discussion, a proper transformation is needed to map \(var^n\) to \(rat^n_s\). The details of the transformation will be given in the following part.

\textbf{Mapping Strategy: } One simple intuition is to map larger \(var^n\) to larger \(rat^n_s\) since input samples that are judged as to be easily classified by the SGNet are expected to bypass more blocks. Generally, a relatively shallow network can correctly classify a large proportion of input samples, indicating that most input samples are ``easy'' samples and only few are hard to be correctly classified. Based on this observation, an exponent function-based mapping strategy is proposed and can be expressed as follows:
\begin{eqnarray}\label{eqn: soft2}
\begin{split}
rat^n_{s} = 1- L^{1- scale\times var^n}/L, \ n \in \{1,2,\cdots,N\}, \quad scale = -\frac{Mln(s_{max})}{ln(L)}
\end{split},
\end{eqnarray}
where \(s_{max}\) denotes the allowed maximum drop ratio and \(rat^n_{s}\in[0, s_{max}]\). \(scale\) transforms \(var\) to the level of difficulties. Considering \(rat^n_{s} \leq s_{max}\) (avoid model with too little complexity) and \(var^n \leq 1/M\), we can get \(scale \times var^n \geq -\frac{Mln(s_{max})}{ln(L)}\).            The proposed mapping strategy tends to map more different \(var^n\) values to large \(rat^n_s\). This approach is consistent with the distribution of the ``easy'' and ``hard'' samples as discussed above. 

At the training phase, the SGAD, which includes the BMNet, the SGNet and the ResNet, can be end-to-end trained from scratch. As shown in Fig.~\ref{fig: SGAD}, Input samples will be fetched to the ResNet and the SGNet simultaneously. The SGNet outputs its own classification results as well as the guideline. The BMNet fetches the first layers's output of the ResNet to produce the binary mask. The ResNet learns to adaptively drop the rest residual blocks based on the output of the binary mask and also produces its own classification results. Then, all the weights in SGAD are updated based on regularization loss \(R^m\), the classification error of the SGNet \(R^g\) and the ResNet \(R^{'}\). The final loss function of SGAD can be expressed as:
\begin{eqnarray}\label{eqn: soft3}
\begin{split}
R_{SGAD}=\alpha R^{'} + \alpha^{m} R^{m} + \alpha^{g} R^{g}
\end{split},
\end{eqnarray}
where \(\alpha\), \(\alpha^{m}\) and \(\alpha^{g}\) denote the weighting factors for ResNet, BMNet and SGNet, respectively. During inference, the regularization term is useless. \textbf{Thus, the SGNet can be removed and only the BMNet and the ResNet are needed after training.}

\section{Experiments}\label{sec: exp}
We evaluate the performance of the proposed SGAD on two datasets: CIFAR-10 and CIFAR-100. The influences of the guideline is investigated. In addition, we also explore different blocks' contributions to the model capacity and the dropping behavior of SGAD. 

\textbf{Model Size}: Both ResNet-32 and ResNet-110 are adopted as baseline in our experiments. The details of structures can be referred to ~\cite{He2016Deep}. The design of BMNet is crucial to the overall complexity of SGAD. On CIFAR-10, the use of BMNets only introduces \(0.06\%\) and \(0.02\%\) computation overheads as compared with ResNet-32 and ResNet-110, respectively. The memory overheads of BMNets are \(1.61\%\) and \(1.59\%\) compared to ResNet-32 and ResNet-110, respectively. Hence, using the proposed BMNet will only render very minor overheads.

\textbf{Training Details}: PyTorch is used to implement the SGAD. The stochastic gradient descent is used as optimizer with momentum 0.9.  The learning rate is initialized at 0.1 and decayed by \(10^{-1}\) after the 128, 160 and 192 epochs. SGAD is trained for 220 epochs with a batch size of 128. \(\alpha, \alpha_{m}\) and \(\alpha_{g}\) are set to 1.0, 1.0 and 0.3 by default, respectively. In our experiments, only adjusting \(s_{max}\) while leaving other hyper-parameters as default can affect the dropping behavior and works in most cases. The last block in SGAD is fixed for all inputs in order to ensure more robust output predictions.

\subsection{Comparisons and Discussion}\label{sec: dis}
We train the SGAD model under two typical settings: 1) a relatively smaller \(s_{max}\), resulting in a model (MF-SGAD) with more FLOPs. 2) a larger \(s_{max}\), which produces a model (LF-SGAD) with less FLOPs. For the latter case, we fine tune the model from a pre-trained MF-SGAD to obtain a faster convergence instead of training from random initialization. The performances of MF-SGAD and LF-SGAD are shown in Table.~\ref{tab: retrain}. For comparison, we also provide the training results of the original ResNets. It can be found that at most cases, for smaller \(s_{max}\), the SGAD can achieve comparable and even better accuracy with less FLOPs as compared to the original ResNets, which indicates the effectiveness of the proposed SGAD. More aggressive reduction in FLOPs can also be obtained under large \(s_{max}\) at a cost of small accuracy loss. For example, the FLOPs can be reduced by 77\(\%\) with only 0.87\(\%\) loss in accuracy (CIFAR10, 110 layers).

%\textbf{Fine Tuning:} To further improve the performance of SGAD, we apply the method of fine tuning and enhance the efficiency of the models. The procedure of training can be decomposed to two stages: 1) a SGAD with a relatively smaller \(s_{max}\) will be trained to get a model with more FLOPs (MF-SGAD). 2) The MF-SGAD is then fine tuned with a higher \(s_{max}\). A model with less FLOPs (LF-SGAD) can be obtained. The performances of MF-SGAD and LF-SGAD are shown in Table.~\ref{tab: retrain}. \textcolor{blue}{For CIFAR-10, the FLOPs can be reduced by more than 63\(\%\) with only 0.5\(\%\) loss in accuracy after fine tuning. For CIFAR-100, the SGAD with fine tuning can reduce the FLOPs by \(19\%\) without any loss in accuracy.}
\begin{table}[htb]
	\centering
	\begin{threeparttable}
		\caption{Results of SGADs and ResNets. For the original ResNets, n-FLOPs=1.0.}
		\label{tab: retrain}
		\begin{tabular}{|c|c||c||c|c||c|c|}
			\hline
			\multirow{2}{*}{Dataset} & \multirow{2}{*}{Layers}& ResNet &\multicolumn{2}{c||}{MF-SGAD, \(s_{max}=0.2\)}   & \multicolumn{2}{c|}{LF-SGAD, \(s_{max}=0.8\)}  \\ \cline{3-7}
			& & accuracy&accuracy  & n-FLOPs &  accuracy & n-FLOPs  \\ \hline
			\multirow{2}{*}{CIFAR-10} & 32 &93.02\(\%\) &93.11\(\%\) & 0.86 &  92.18\(\%\) & 0.47 \\
			 & 110 & 94.57\(\%\)&94.20\(\%\) & 0.86 &  \textbf{93.70\(\%\)} & \textbf{0.23} \\ \hline
			\multirow{2}{*}{CIFAR-100} & 32 & 70.38\(\%\)& 70.85\(\%\) & 0.77 &  70.09\(\%\) & 0.71 \\
			 & 110 &73.94\(\%\) & 73.94\(\%\) & 0.94 &  73.94\(\%\) & 0.75  \\ \hline
			
		\end{tabular}	

	\end{threeparttable}
\end{table}

\begin{figure} [hbt]
	\centering
	\includegraphics[width=5.4in]{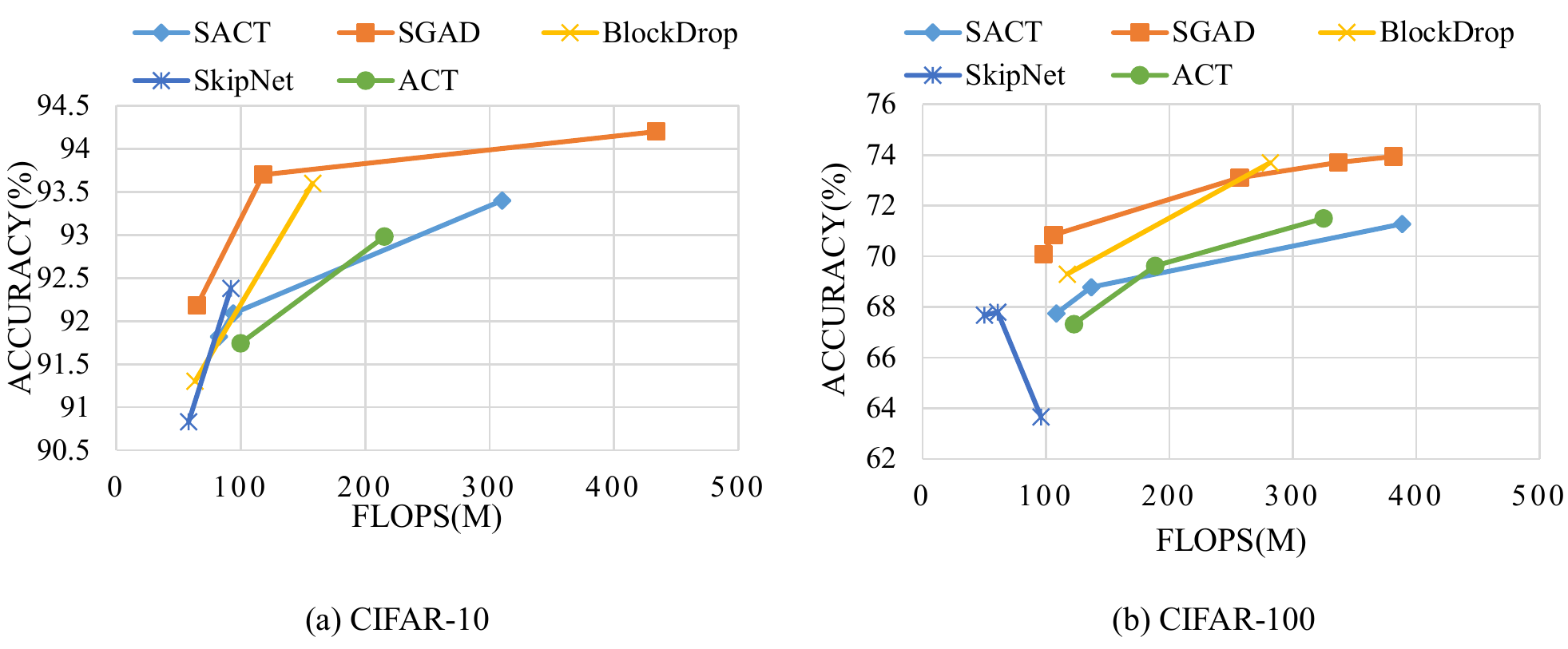}
	\caption{Comparisons with state-of-the-art. (a) Results on CIFAR-10. (b) Results on CIFAR-100. The solid lines and dashed lines have different baselines.}\label{fig: comp}
\end{figure}

\textbf{Comparisons with Existing Works:} In this subsection, we compare the proposed SGAD with previous works. The performances are shown in Fig.~\ref{fig: comp}, which contains the results of SACT~\cite{figurnov2016spatially}, ACT~\cite{figurnov2016spatially}, SkipNet~\cite{wang2017skipnet}, and BlockDrop~\cite{wu2017blockdrop}. The proposed SGAD outperforms all existing networks at most cases. 

\footnotetext[2]{\url{https://github.com/mfigurnov/sact}}

For ACT and SACT, since the results on CIFAR are not reported, we conduct the experiments using the code$^\dag$ provided by the authors of SACT. Compared with the SACT, the FLOPs of SGAD can be reduced by \(70\%\) with even \(0.3\%\) higher accuracy on CIFAR-10. On CIFAR-100, the accuracy can be enhanced by \(1.8\%\) with \(34\%\) less FLOPs. The proposed SGAD also outperforms other algorithms such as ACT and SkipNet. Compared with the BlockDrop which currently achieves the state-of-the-art results, SGAD can also improve the accuracy by \(0.1\%\) with \(25\%\) less computational complexity on CIFAR-10.

\begin{figure} [hbt]
	\centering
	\includegraphics[width=5.4in]{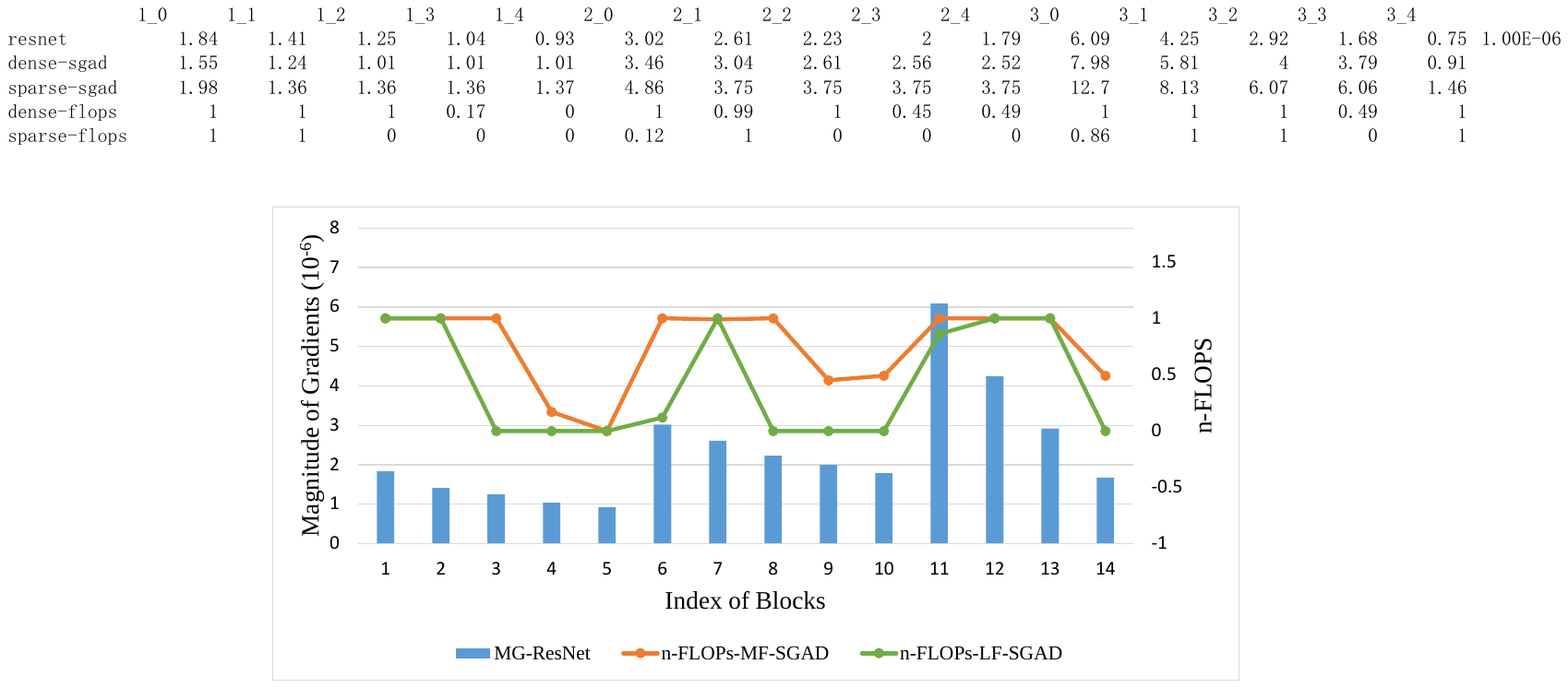}
	\caption{Comparisons of magnitude of gradients and normalized flops of each blocks. The n-FLOPS-MF-SGAD and the n-FLOPS-LF-SGAD denote the n-FLOPS of MF-SGADs and LF-SGAD, respectively. The magnitudes of gradients are given by the mean value of L1 normalization of gradients after the \(160\)-th epoch. }\label{fig: dis}
\end{figure}

\textbf{Discussion: }{The dropping behavior of SGAD is explored here. The experiments are conducted using ResNet-32 and SGAD-32 on CIFAR-10. Fig.~\ref{fig: dis}  shows the comparisons of magnitude of gradients and normalized flops of each blocks. Since the last block is always fixed in SGAD, the n-FLOPs of the last block is always 1 and is not listed. It is worth noting that in ResNet32, every 5 blocks share the same number of output channels, \textbf{C-block} is used here to denote a cluster of 5 blocks. From Fig.~\ref{fig: dis} we can obtain the followings:
\begin{enumerate}
	\item In the original ResNets, different blocks usually have different magnitudes of gradients (MGs). In each C-block, the MGs decrease gradually from the first block to the fifth block. Such phenomenon shows that the first several blocks in a C-block have relatively higher LCMC than the others. The discovery gained here is consistent with the reports from previous works~\cite{veit2016residual,jastrzebski2017residual}.
	\item According to our experiments, in each C-block, the dropping behavior is closely related to the MGs. Blocks with higher MGs usually have a higher n-FLOPs. Combining the analysis in Section~\ref{sec: mask} and the experimental results, the blocks with less \emph{MGs} are tended to be skipped. Thus, the updates of these blocks are further decreased in SGAD. To reduce the FLOPs while maintaining the performance, SGAD tries to keep the blocks with higher \emph{LCMC}.
	\item  As shown in Fig.~\ref{fig: dis}, some blocks will be skipped by all the input samples after training, thus leads to zero n-FLOPs for these blocks. As an additional benefit, these \emph{dead blocks} can be removed during inference to reduce the memory storage requirement.
\end{enumerate}

\section{Conclusion and Future Work}
SGAD is proposed to exploit an adaptive processing pattern for different input samples. To enable the propagation of gradients, STE is introduced to approximate the non-differential rounding function during the training phase. The information contained in softmax layer is explored to inform the SGAD the difficulties of various input samples being classified correctly. In addition, a dedicatedly designed mapping strategy is introduced to combine the difficulties and the dropping ratio. The experiments demonstrate that the proposed SGAD outperforms previous works under the same baselines. While the reduction in FLOPs may not accurately reflect the real running latencies under different hardware devices (\textit{eg.} CPUS, GPUs), the real speedup measurements will be conducted in the future.

\bibliographystyle{named}
\bibliography{ijcai18}

\end{document}